\def\paperID{11651}
\def\confName{CVPR}
\def\confYear{2024}

\def\paperTitle{C-BEV: Contrastive Bird's Eye View Training for Cross-View Image Retrieval and 3-DoF Pose Estimation}

\def\authorBlock{
	\newcommand{\spacing}{\hspace{0.3mm}}
	\hspace{-3mm}
	Florian Fervers$^1$ \spacing
	Sebastian Bullinger$^1$ \spacing
	Christoph Bodensteiner$^1$ \spacing
	Michael Arens$^1$ \spacing
	Rainer Stiefelhagen$^2$ \\
	$^1$Fraunhofer IOSB \quad $^2$Karlsruhe Institute of Technology \\
	{\tt\small $^1$\{firstname.lastname\}@iosb.fraunhofer.de \quad $^2$rainer.stiefelhagen@kit.edu}
}

\newif\ifreview 
\newif\ifarxiv \newcommand{\arxiv}{\arxivtrue}
\newif\ifcamera 
\newif\ifrebuttal 

\arxiv %

\pdfoutput=1
\documentclass[10pt,twocolumn,letterpaper]{article}
\ifreview \usepackage[review]{cvpr} \fi
\ifarxiv \usepackage[pagenumbers]{cvpr} \fi
\ifrebuttal \usepackage[rebuttal]{cvpr} \fi
\ifcamera \usepackage{cvpr} \fi

\usepackage{graphicx}	
\usepackage{amsmath}	
\usepackage{amssymb}	
\usepackage{booktabs}
\usepackage{times}
\usepackage{microtype}
\usepackage{epsfig}
\usepackage[table,xcdraw,dvipsnames]{xcolor}
\usepackage{caption}
\usepackage{float}
\usepackage{placeins}
\usepackage{color, colortbl}
\usepackage{stfloats}
\usepackage{enumitem}
\usepackage{tabularx}
\usepackage{xstring}
\usepackage{multirow}
\usepackage{xspace}
\usepackage{url}
\usepackage{subcaption}
\usepackage{xcolor}
\usepackage[hang,flushmargin]{footmisc}

\ifcamera \usepackage[accsupp]{axessibility} \fi

\ifarxiv  \fi

\newcommand{\R}[1]{{%
    \textbf{%
        \ifstrequal{#1}{1}{\textcolor{red}{R#1}}{%
        \ifstrequal{#1}{2}{\textcolor{blue}{R#1}}{%
        \ifstrequal{#1}{3}{\textcolor{magenta}{R#1}}{%
        \ifstrequal{#1}{4}{\textcolor{teal}{R#1}}{%
                           \textcolor{cyan}{R#1}%
        }}}}%
    }%
}}

\usepackage{xr-hyper}

\makeatletter
\newcommand*{\addFileDependency}[1]{
  \typeout{(#1)}
  \@addtofilelist{#1}
  \IfFileExists{#1}{}{\typeout{No file #1.}}
}

\makeatother
\newcommand*{\myexternaldocument}[1]{
    \externaldocument{#1}
    \addFileDependency{#1.tex}
    \addFileDependency{#1.aux}
}

\definecolor{cvprblue}{rgb}{0.21,0.49,0.74}
\usepackage[pagebackref,breaklinks,colorlinks,citecolor=cvprblue]{hyperref}
\usepackage[capitalize]{cleveref}
\crefname{section}{Sec.}{Secs.}
\crefname{table}{Table}{Tables}
\crefname{figure}{Fig.}{Figs.}

\ifarxiv \crefname{appendix}{App.}{Apps.}
\else \crefname{appendix}{Suppl.}{Suppls.} \fi

\unless\ifarxiv \myexternaldocument{_supplementary} \fi

\usepackage{amssymb}
\usepackage{pifont}
\usepackage{float}
\usepackage{capt-of,etoolbox}

\usepackage{xcolor}
\usepackage{arydshln}
\usepackage{hhline}

\usepackage[free-standing-units=true]{siunitx}

\makeatletter
\renewcommand{\paragraph}{%
	\@startsection{paragraph}{4}%
	{\z@}{0.7mm}{-1em}%
	{\normalfont\normalsize\bfseries}%
}
\makeatother

\begin{document}
\title{\paperTitle}
\author{\authorBlock}
\maketitle

\begin{abstract}

To find the geolocation of a street-view image, \mbox{cross-view} geolocalization (CVGL) methods typically perform image retrieval on a database of georeferenced aerial images and determine the location from the visually most similar match. Recent approaches focus mainly on settings where \mbox{street-view} and aerial images are preselected to align \wrt translation or orientation, but struggle in challenging \mbox{real-world} scenarios where varying camera poses have to be matched to the same aerial image. We propose a novel trainable retrieval architecture that uses \mbox{bird's eye view (BEV)} maps rather than vectors as embedding representation, and explicitly addresses the \mbox{many-to-one} ambiguity that arises in \mbox{real-world} scenarios. The BEV-based retrieval is trained using the same contrastive setting and loss as \mbox{classical retrieval}.

Our method C-BEV surpasses the state-of-the-art on the retrieval task on multiple datasets by a large margin. It is particularly effective in challenging many-to-one scenarios, \eg increasing the top-1 recall on VIGOR's cross-area split with unknown orientation \mbox{from 31.1\% to 65.0\%}. Although the model is supervised only through a contrastive objective applied on image pairings, it additionally learns to infer the \mbox{3-DoF} camera pose on the matching aerial image, and even yields a lower mean pose error than recent methods that are explicitly trained with metric groundtruth.

\end{abstract}
\section{Introduction}
\label{sec:intro}

\begin{figure}[t]
	\centering
	\includegraphics[width=\linewidth]{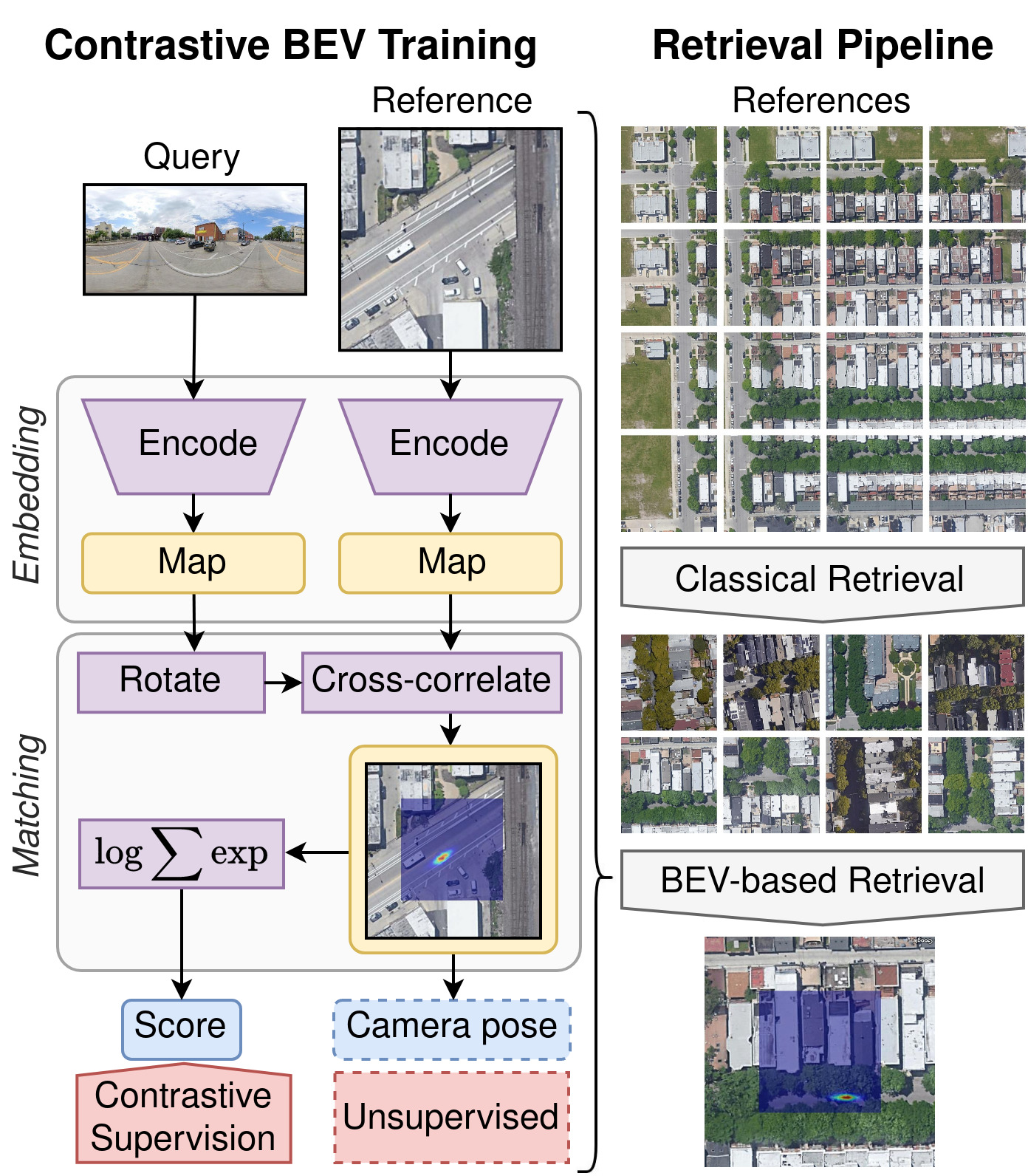}
	\vspace{-6mm}
	\caption{Overview of our novel cross-view retrieval method that uses bird's eye view (BEV) maps as embedding representation, and finds a match by testing different 3-DoF camera poses of the \mbox{street-view} query on the aerial image. The model is trained using the same loss as classical retrieval architectures (\eg CLIP \cite{radford2021learning}), and additionally learns to infer the 3-DoF camera pose on the aerial image without explicit supervision. At test time we employ a \mbox{two-stage} approach where candidates for a given query are determined using classical retrieval and reranked using BEV-based retrieval.}
	\label{fig:overview}
	\vspace{-4mm}
\end{figure}

Visual geolocalization refers to the task of finding the geolocation of a street-view image by matching it with a database of geolocalized reference images. In the same-view setting, the query image is captured from a similar perspective as the reference images, \eg using Google Street View \cite{googlestreetview}. Such data provide similar visual cues that facilitate the matching process, but also limit applicability to regions where they are available and sufficiently up-to-date. In the cross-view setting - also called cross-view geolocalization (CVGL) - the street-view query image is instead compared against aerial imagery as reference data. While this requires the matching to account for drastic changes in viewpoint and scale, it allows for large-scale application due to the global availability of georegistered orthophotos \cite{googlemaps}.

Current research addresses this task by tiling the aerial imagery into smaller patches and retrieving the most similar aerial image for a given street-view query image \cite{zhu2022transgeo,deuser2023sample4geo}. The similarity is determined by training a model to map \mbox{street-view} and aerial images into a joint embedding space using a contrastive objective. The corresponding embedding distance reflects the likelihood that a query image is located on a reference aerial image, or a search region in the center of the image.

\begin{figure}
	\centering
	\begin{subfigure}{0.335\columnwidth}
		\centering
		\includegraphics[height=55.0mm]{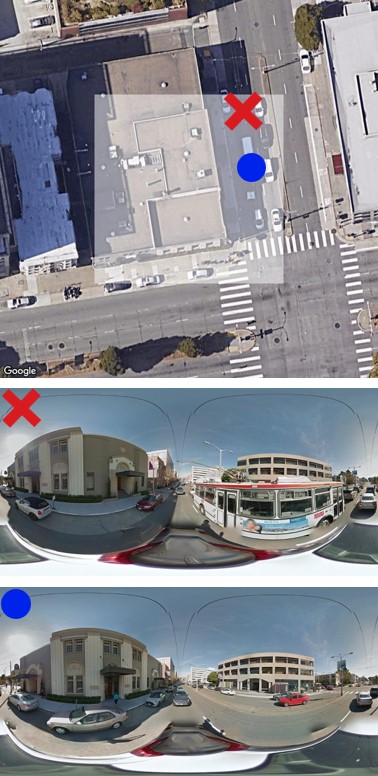}
		\caption{Similar street-views.}
	\end{subfigure}%
	\hspace{0.45mm}%
	\begin{subfigure}{0.655\columnwidth}
		\centering
		\includegraphics[height=55.0mm]{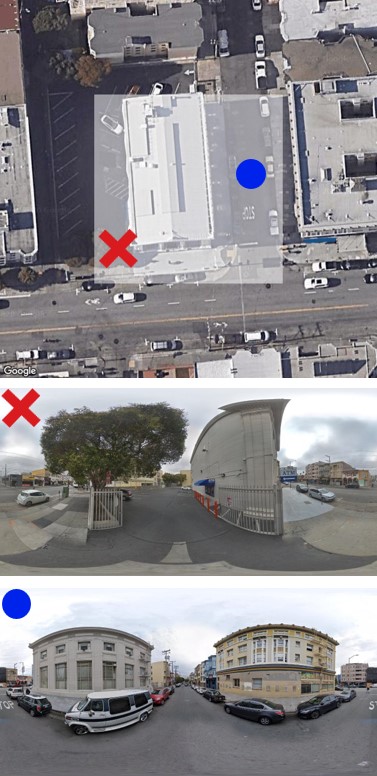}%
		\hspace{1mm}\includegraphics[height=55.0mm]{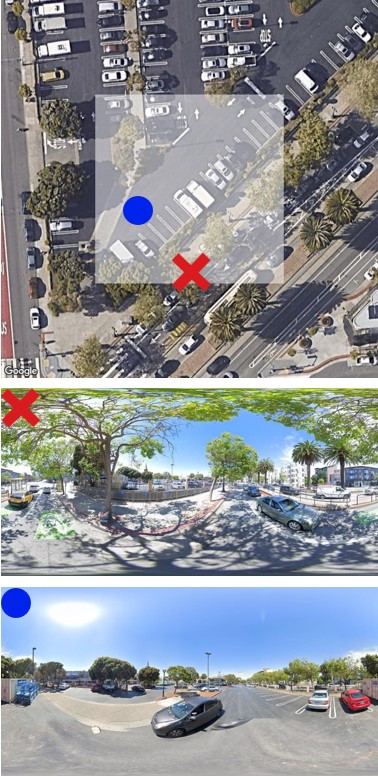}
		\caption{Dissimilar street-views.}
	\end{subfigure}
	
	\vspace{-2mm}\caption{Examples of aerial images with paired street-view images in the VIGOR \cite{zhu2021vigor} dataset. Different camera poses potentially lead to vastly differing street-view appearance, but have to be matched to the same aerial image. Search regions are shown in white.}
	\label{fig:simdissim}
	\vspace{-3mm}
\end{figure}

Embeddings have to capture relevant information from the input images while being invariant to non-distinctive image characteristics (\eg lighting conditions). This presents a particular challenge in cross-view image retrieval: Different camera poses within the aerial image's search region may lead to vastly differing street-view appearances and little to no overlap between the camera images, but have to be mapped to vector representations that align with the same aerial image (\cf \cref{fig:simdissim}).

To address this problem, we digress from the established approach of using vector embeddings, and instead utilize \mbox{bird's eye view (BEV)} maps as embedding representation. A BEV map is given as a two-dimensional feature map centered on the 3-DoF camera pose. This removes the requirement for the model to predict pose-invariant embeddings, and instead shifts resolving the pose ambiguity to the matching operation between embeddings.

We test for a match between a query and reference image by considering a set of possible 3-DoF camera poses in the aerial image's search region and aggregating the individual pose likelihoods. Each camera pose is tested by comparing the BEV map with the aerial image under the respective transformation. The entire model is end-to-end differentiable up to the retrieval scores and is trained using the same contrastive supervision as classical vector-based architectures. At test time, we first employ vector embeddings to retrieve a set of candidates and prior probabilities, and perform reranking in a second step using BEV embeddings.

Our method C-BEV achieves higher recall than previous works and is particularly effective in challenging real-world scenarios where the street-view and aerial images are not preselected to align \wrt translation and orientation. Additionally, the model implicitly learns to infer the metric 3-DoF pose of the street-view camera on the matching aerial image (\cf \cref{fig:probs}) \textit{without seeing metric groundtruth during training}. The explicit geometric constraints in the BEV map and the matching operator represent a strong inductive bias that condition the model to solve the retrieval task by providing an estimate of the metric camera pose. The only training supervision is given via the pairwise assignment of street-view and aerial images per batch.

To summarize, our main contributions are as follows:
\begin{enumerate}
	\item We introduce our approach C-BEV for the cross-view retrieval task that uses a novel embedding representation in the form of BEV maps and specifically addresses the many-to-one ambiguity between street-view and aerial images that arises in real-world scenarios.
	\item We show that our model is able to infer the \mbox{3-DoF} camera pose on the matching aerial image despite only being supervised through the retrieval objective, \ie pairwise assignment of street-view and aerial images.
	\item We perform extensive evaluation on several datasets to demonstrate that (a) C-BEV surpasses \mbox{state-of-the-art} methods on the retrieval task by a large margin and is particularly effective in more challenging scenarios, and (b) C-BEV yields 3-DoF camera poses that compare favorably even with recent methods that are explicitly supervised to perform \mbox{3-DoF} camera pose estimation.
\end{enumerate}

\section{Related work}
\label{sec:related}

Cross-view geolocalization refers to the problem of finding the geolocation of a street-view image by matching it against georeferenced aerial images. While early approaches use handcrafted features to match images from both domains \cite{bansal2011geo,viswanathan2014vision}, to our best knowledge Lin \etal \cite{lin2015learning} and Workman \etal \cite{workman2015wide} present the first works that learn an image embedding to address the task as a retrieval problem. Workman \etal use a pretrained model for the street-view images and train a second model to predict aligning embeddings for the aerial images. Lin \etal utilize a contrastive learning objective to train a siamese model for street-view and aerial images, followed by a long thread of works that expand on this idea \cite{vo2016localizing,hu2018cvm,liu2019lending,cai2019ground,shi2019spatial,zhu2021vigor,zhu2022transgeo,deuser2023sample4geo}.

\paragraph{Architecture} Several dedicated architectures have been proposed to extract embedding vectors from the input images \cite{shi2019spatial,shi2020optimal,yang2021cross,zhu2023simple,zhang2023cross}. Many works use modules that implement a global receptive field per image, \eg through spatial attention pooling \cite{shi2019spatial,zhang2023cross} or a multilayer perceptron (MLP) along the spatial dimensions \cite{zhu2023simple}. This addresses the limited receptive field of convolutional neural networks (CNN) and helps the network reason about the spatial layout of the scene. Recently, the transformer architecture \cite{vaswani2017attention,dosovitskiy2020image} has been applied to CVGL, either through a stand-alone vision transformer (ViT) \cite{zhu2022transgeo} or in combination with convolutional neural networks (CNN) \cite{yang2021cross,zhu2023simple}.

\paragraph{Loss} Several recent works employ the infoNCE \cite{oord2018representation} loss rather than a previously used (soft margin) triplet loss to model the contrastive learning objective \cite{zhu2023simple,deuser2023sample4geo}, which has been successfully employed in other domains such as for contrastive language-image pretraining (CLIP) \cite{radford2021learning}. Additional auxiliary loss functions have been proposed to aid the training process, such as regressing the relative rotation \cite{vo2016localizing,cai2019ground} and metric translation \cite{zhu2021vigor} between matching aerial and ground images, or using the model as a generative adversarial network (GAN) to generate realistic aerial views from ground images \cite{regmi2019bridging} or ground views from aerial images \cite{toker2021coming}.

\paragraph{Global hard sample mining} As the training progresses, the model tends to classify most samples in a batch correctly and thus receives little to no supervision from the batch. To address this issue, several works employ a global hard sample mining strategy: During training, the embedding features for all images are stored in memory and updated either after every training batch \cite{zhu2021vigor,zhu2021revisiting,zhu2022transgeo}, or by running inference on the entire training split every few epochs \cite{deuser2023sample4geo}. The features are used to globally mine hard samples for every training batch.

\paragraph{Polar transform} Some works \cite{shi2019spatial,shi2020looking,toker2021coming} warp the aerial image into a polar-transformed view around the position of the ground camera, such that it more closely resembles the perspective of a street-view panorama. This is only applicable on benchmarks where aerial and street-view images are preselected such that the camera's position on the aerial image is known in advance.

\paragraph{State-of-the-art} Deuser \etal \cite{deuser2023sample4geo} demonstrate that an \mbox{off-the-shelf} architecture without domain-specific adjustments is sufficient to outperform previous works by a large margin. They train a modern CNN \cite{liu2022convnet} with a large batch size, global hard sample mining and suitable hyperparameters without using \mbox{polar-transform}, specialized modules or auxiliary loss functions, and achieve state-of-the-art results.

We use a retrieval architecture based on vector embeddings following their approach to yield a set of candidates for a given street-view query. In a second step, we utilize our novel BEV embeddings to rerank these candidates and achieve significantly improved recall.

\paragraph{Metric CVGL} While cross-view image retrieval is able to account for large (\eg city-scale) search regions efficiently, it provides rough geolocations only up to the size of the search region per aerial image. The metric CVGL task aims to estimate the relative 3-DoF camera pose within a matching aerial image's search region. This problem has recently gained attention in the research community \cite{xia2021cross, xia2022visual, zhu2021vigor,fervers2022continuous,shi2022beyond}, with several works employing BEV maps that are matched with the aerial image over a set of discrete camera poses \cite{fervers2023uncertainty,sarlin2023orienternet,sarlin2023snap,shi2023boosting}.

These methods require an explicit supervision of the camera pose with accurate groundtruth that is often hard to obtain and requires refinement steps (\eg pseudo-labels \cite{fervers2023uncertainty} or structure-from-motion \cite{sarlin2023orienternet}) that represent restrictions on what type of data can be used. In contrast, our method learns to infer the 3-DoF camera pose despite only being supervised through a contrastive objective based on image pairings, and thus requires no metric groundtruth for the training.

\section{Method}
\label{sec:method}

\subsection{Problem formulation}\label{sec:problem-formulation}

We consider a dataset with $n_s$ street-view images and $n_a$ aerial images. Each aerial image covers a square search region in its center (\cf \cref{fig:simdissim}). If the camera location of a street-view image is within the search region, it is defined as a match for the respective aerial image. The search regions of all reference images are disjoint and cover all query locations, such that each street-view image matches with exactly one aerial image. The objective is to retrieve the matching aerial reference image for a given street-view query, and thereby provide an estimate of its geolocation.

The \mbox{per-image} search regions are typically chosen to cover a (\eg city-scale) target region densely (e.g. VIGOR dataset \cite{zhu2021vigor}). This constitutes a many-to-one matching problem since street-view images from various camera poses have to be matched with the same aerial image (\cf \cref{fig:simdissim}). Earlier datasets represent a one-to-one mapping instead where an aerial image is preselected for each street-view query to be centered on its location (e.g. CVUSA \cite{workman2015wide}, CVACT \cite{liu2019lending}). While street-view panoramas and aerial images in VIGOR, CVUSA and CVACT are north-aligned, we also consider the case where the street-view orientation relative to the aerial image is unknown.

Given a match between a pair of street-view and aerial images, the metric CVGL task aims to estimate the metric 3-DoF pose of the street-view image in the search region of the respective aerial image.

\subsection{Overview}

We present a two-stage approach to address the CVGL task (\cf \cref{fig:overview}):\vspace{-1mm}
\begin{description}
	\item[Stage 1] Use vector embeddings to retrieve a set of possible matches from the database for a given street-view query. This allows for a cheap comparison and search in the database.\vspace{-2mm}
	\item[Stage 2] Use BEV embeddings to jointly (a) perform reranking on the set of candidates and (b) predict the metric 3-DoF camera pose on matching aerial images.\vspace{-1mm}
\end{description}

The first stage is responsible for ruling out easy negatives, \eg when a discrimination is possible based mainly on image semantics without considering the specific camera pose on the aerial image. It allows for a cheap query of the database especially when combined with efficient search structures. The second stage has a stronger geometric bias and explicitly considers varying camera poses to improve the recall and perform 3-DoF pose estimation (\ie 2D translation and 1D rotation). Since it is computationally more expensive than the first stage and does not utilize efficient search structures, it is only applied on the set of retrieved candidates. The final score for a given reference image is computed from the vector-based prior and BEV-based update.

We separately train a model for each stage using the same contrastive image retrieval objective. At test time, the models are applied consecutively to retrieve matches for a given street-view query and predict the 3-DoF camera pose.

\subsection{Stage 1: Retrieval with vector embeddings}\label{sec:eval-stage1}

We train a baseline model in an image retrieval setting similar to Deuser \etal \cite{deuser2023sample4geo} to retrieve a set of candidates from the reference database. The model learns to map both street-view and aerial images into a joint embedding space that can be used to perform nearest neighbor search at test time for a given query image.

\paragraph{Model} We choose the ConvNeXt \cite{liu2022convnet} architecture as backbone for the model. The output features are \mbox{mean-pooled} and \mbox{L2-normalized} to produce the embedding vector for a given image. We use a single model for both \mbox{street-view} and aerial domains which has been shown to yield better results than training separate domain-specific models \cite{deuser2023sample4geo}.

\paragraph{Training} The model is trained in a contrastive setting similar to CLIP \cite{radford2021learning}. For each training batch we sample $n$ pairs of matching \mbox{street-view} and aerial images, such that each batch represents a one-to-one mapping even if the whole dataset represents a many-to-one problem. We use a symmetric cross-entropy loss that maximizes the cosine similarity of the $n$ correct pairings, and minimizes the similarity of the $n^2 - n$ incorrect pairings in the batch.

Since the model tends to classify most samples per batch correctly as the training progresses, we employ a global hard sample mining strategy to provide stronger supervision per batch. Every $e$ epochs, the embedding features of all training images are computed and stored in memory. For each batch, we sample hard negatives using the stored embeddings following the strategy introduced by Deuser \etal \cite{deuser2023sample4geo}. During the first $e$ epochs, the haversine distance is used for determining hard negatives.

\subsection{Stage 2: Retrieval with BEV embeddings}

In the second stage, we utilize BEV maps rather than vectors as embedding representation that are compared to determine the matching score for a pair of images (\cf \cref{fig:overview}). The BEV map is given in local 3-DoF camera coordinates, and thus mitigates the challenge for the model to predict embeddings for varying street-view camera poses that directly align with the aerial image embedding.

For each aerial image, a set of possible camera poses is sampled in a grid over the 3-DoF search volume and tested by comparing the street-view BEV map with the aerial image under the respective transformations \cite{fervers2023uncertainty,sarlin2023orienternet}. The matching score for an image pair is determined via the sum of all pose probabilities and trained end-to-end using the same loss as in \mbox{vector-based} retrieval. Due to the strong geometric bias in the architecture, the model learns to infer the correct 3-DoF camera pose despite receiving no direct pose supervision through the use of metric groundtruth.

Given the higher computational cost of predicting and comparing BEV maps, we apply the second stage to a set of candidates retrieved using the first stage rather than the entire dataset.

\begin{figure}
	\centering
	\includegraphics[width=\linewidth]{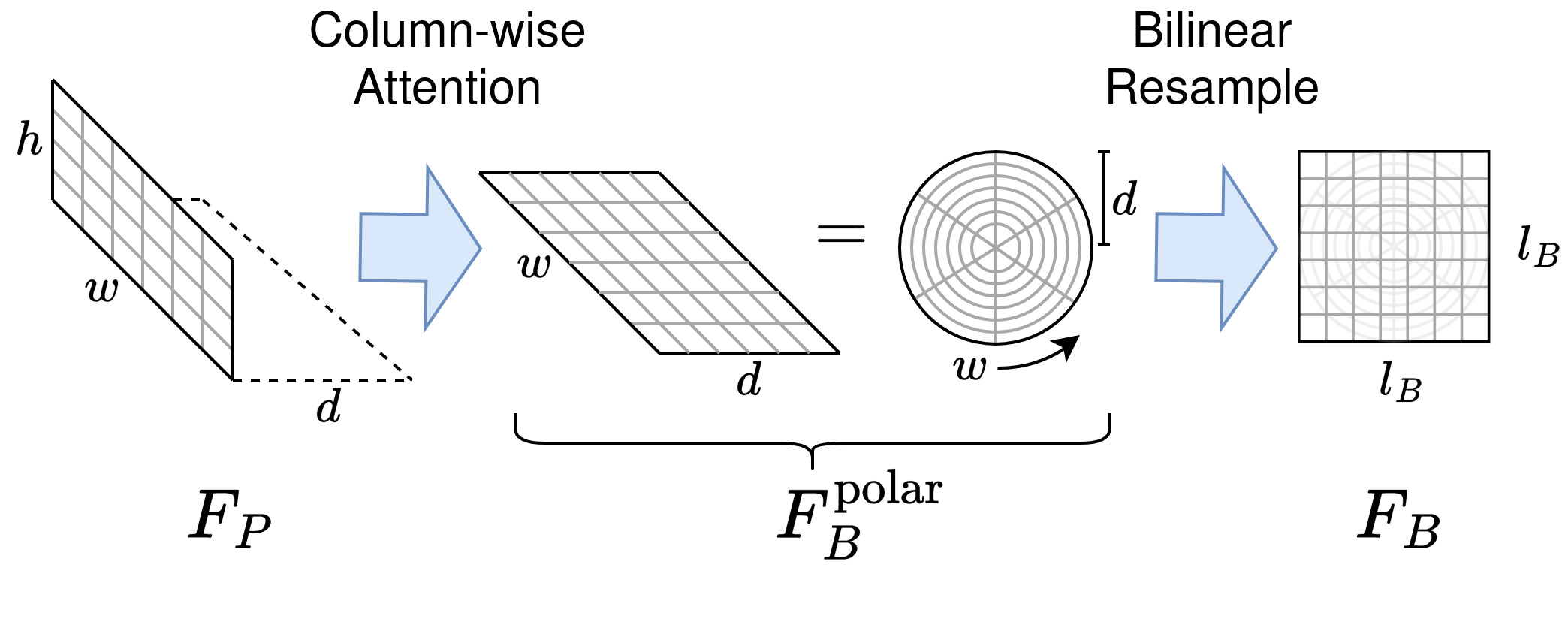}
	\vspace{-9mm}
	\caption{Transformation from street-view panorama to bird's eye view (BEV). The feature map of the panorama image $F_P$ with height $h$ and width $w$ is transformed into a polar BEV representation $F_B^{\textit{polar}}$ with $w$ angles and $d$ depths using an attention operation along each column. The features are then resampled bilinearly from polar to cartesian coordinates $F_B$.}
	\label{fig:pano2bev}
	\vspace{-4mm}
\end{figure}

\paragraph{BEV encoder} The embedding maps in BEV are inferred from the street-view and aerial input images. In the aerial domain, the image is already given in BEV with a known resolution and is transformed into a feature map \mbox{$F_A \in \mathbb{R}^{l_A \times l_A \times c}$} with sidelength $l_A$ and $c$ channels using a fully-convolutional network $N_A$. In the street-view domain, the image is given as a panorama and transformed into a BEV map \mbox{$F_B \in \mathbb{R}^{l_B \times l_B \times c}$} using a second model $N_B$ as described below (\cf \cref{fig:pano2bev}) following related works \cite{philion2020lift,roddick2020predicting,saha2022translating,sarlin2023orienternet}.

We first extract a feature map $F_P$ from the \mbox{street-view} panorama with height $h$ and width $w$ using a \mbox{fully-convolutional} network. We then transform $F_P$ into a BEV representation $F_B^{\textit{polar}}$ in polar coordinates with $w$ angles and $d$ depths using an attention operation: Each cell in $F_B^{\textit{polar}}$ computes a weighted average over the corresponding column in $F_P$. The attention weights are predicted from $F_P$ as a categorical distribution over $d$ depths and normalized along the vertical axis using a softmax operation. Finally, $F_B^{\textit{polar}}$ is resampled bilinearly from polar to cartesian representation $F_B$. We maintain a mask with the same resolution that indicates valid cells in $F_B$ up to a distance of $\frac{1}{2}l_B$ to the camera. Both $F_A$ and $F_B$ are L2-normalized globally before applying the matching operation.

\paragraph{BEV matching} We sample a set of discrete 3-DoF camera pose hypotheses $\mathcal{H} \subset \text{SE(2)}$ in a regular grid over the search region of the aerial image with $n_t$ translations per dimension and $n_\theta$ orientations. Each pose hypothesis $p \in \mathcal{H}$ is tested by transforming $F_B$ into the coordinate system of $F_A$ using $p$ and computing the inner product between both maps, yielding an unnormalized logit for $p$. The results for all poses are stored in a $n_t \times n_t \times n_\theta$ score volume $S$. \cite{fervers2023uncertainty}

The entire volume $S$ can be evaluated jointly and efficiently by rotating $F_B$ by $n_\theta$ angles and computing the cross-correlation with $F_A$ in the Fourier domain \cite{barnes2019masking,barsan2020learning}. We choose the size of $F_B$ such that all unmasked cells entirely fit onto the aerial map $F_A$ at maximum relative offset:
\begin{equation}\label{eq:lalb}
	n_t \le l_A - l_B + 1
\end{equation}

\paragraph{Retrieval score}

The predicted probability for a match between a pair of images is defined as the the area under the predicted probability density function (PDF) in the aerial image's search region. Since the PDF is represented by unnormalized logits in $S$, we compute the overall score $s \in \mathbb{R}$ for a match as follows:
\begin{equation}\label{eq:lse}
	s = \text{lse}(S) = \log \sum_{p \in \mathcal{H}} \exp S(h)
\end{equation}
 For a given street-view query, the probability distribution over possible aerial reference images is computed by applying a softmax operation over the respective retrieval scores.

\paragraph{Pose estimation} The score volume $S$ represents the likelihoods of individual camera poses in unnormalized logit representation and is transformed into a probability distribution per image using a softmax operation. The final camera pose is computed as the weighted sum over the pose hypotheses using their respective probabilities.

\paragraph{Training} The model is trained end-to-end using the same contrastive setting as vector-based architectures. For a batch of $n$ aerial and $n$ street-view images, we compute the matching scores for $n^2$ pairs and apply the symmetric \mbox{cross-entropy} loss. The score volume $S$ is not supervised directly, but only through the scalar retrieval score $s$.

We employ a global hard sample mining strategy by choosing the hardest neighbors for a single street-view image per batch. Since determining nearest neighbors over the entire dataset using BEV embeddings at training time is computationally infeasible, we instead utilize the learned vector embeddings from the first stage to find hard samples on the training split in the second stage. This allows for an efficient global hard sample mining and conditions the model to specialize on the failure cases of the first stage.

\paragraph{Testing} To find matches for a given street-view query at test time, we employ learned vector embeddings to retrieve the $k$ highest ranked reference images. We add the \mbox{BEV-based} retrieval scores for all candidates onto the prior \mbox{vector-based} scores according to Bayes' Theorem, and rerank the candidates \wrt the final scores.

\section{Experiments}
\label{sec:evaluation}

\begin{table*}[!h]
	\setlength{\tabcolsep}{6pt} %
	\newcommand{\spacing}{\hspace{3mm}}
	\def\arraystretch{0.9}
	\small
	\centering
	\caption{Top-k recall on the VIGOR dataset (\ie with unknown translation between street-view camera and aerial image). Our baseline is the first stage of our method and trained similar to Sample4Geo \cite{deuser2023sample4geo}.}
	\label{tab:vigor-recall}
	\vspace{-3mm}
	\begin{tabular}{l|cccc|cccc}
		& \multicolumn{4}{c|}{Same-area} & \multicolumn{4}{c}{Cross-area} \\
		& R@1 & R@5 & R@10 & R@1\% & R@1 & R@5 & R@10 & R@1\% \\
		\hline

		\textbf{Known orientation} & & & & & & & & \\
		\spacing SAFA \cite{shi2019spatial} & 33.93 & 58.42 & 68.12 & 98.24 & 8.20 & 19.59 & 26.36 & 77.61 \\
		\spacing TransGeo \cite{zhu2022transgeo} & 61.48 & 87.54 & 91.88 & 99.56 & 18.99 & 38.24 & 46.91 & 88.94 \\
		\spacing SAIG \cite{zhu2023simple} & 65.23 & 88.08 & - & \textbf{99.68} & 33.05 & 55.94 & - & 94.64 \\
		\spacing Sample4Geo \cite{deuser2023sample4geo} & 77.86 & 95.66 & 97.21 & 99.61 & 61.70 & 83.50 & 88.00 & \textbf{98.17} \\
		\spacing Baseline & 78.25 & 95.70 & 97.09 & 99.48 & 61.26 & 82.49 & 86.89 & 98.02 \\
		\spacing C-BEV (ours) & \textbf{87.49} & \textbf{97.65} & \textbf{98.26} & 99.48 & \textbf{80.01} & \textbf{92.02} & \textbf{93.33} & 98.02 \\

		\textbf{Unknown orientation} & & & & & & & & \\
		\spacing TransGeo \cite{zhu2022transgeo} & 47.69 & 79.77 & 86.36 & \textbf{99.29} & 5.54 & 14.22 & 19.63 & 66.93 \\
		\spacing Baseline & 66.09 & 90.73 & 93.38 & 98.31 & 31.14 & 52.11 & 59.41 & \textbf{89.14} \\
		\spacing C-BEV (ours) & \textbf{82.60} & \textbf{95.35} & \textbf{96.11} & 98.31 & \textbf{65.01} & \textbf{75.76} & \textbf{76.98} & \textbf{89.14} \\
	\end{tabular}
\end{table*}

\begin{table*}[!h]
	\setlength{\tabcolsep}{4pt} %
	\newcommand{\spacing}{\hspace{3mm}}
	\def\arraystretch{0.9}
	\small
	\centering
	\caption{Top-k recall on the CVUSA and CVACT datasets (\ie with aerial images preselected to be centered on the street-view camera locations). $\dagger$ denotes methods that use polar transformation on the aerial input image. Our baseline is the first stage of our method and trained similar to Sample4Geo \cite{deuser2023sample4geo}.}
	\label{tab:cv-recall}
	\vspace{-3mm}
	\begin{tabular}{l|cccc|cccc|cccc}
		& \multicolumn{4}{c|}{CVUSA} & \multicolumn{4}{c|}{CVACT-Val} & \multicolumn{4}{c}{CVACT-Test} \\
		& R@1 & R@5 & R@10 & R@1\% & R@1 & R@5 & R@10 & R@1\% & R@1 & R@5 & R@10 & R@1\% \\
		\hline
		
		\textbf{Known orientation} & & & & & & & & & & & & \\
		\spacing SAFA$^\dagger$ \cite{shi2019spatial} & 89.84 & 96.93 & 98.14 & 99.64 & 81.03 & 92.80 & 94.84 & 98.17 & - & - & - & - \\
		\spacing DSM \cite{shi2020looking} & 91.96 & 97.50 & 98.54 & 99.67 & 82.49 & 92.44 & 93.99 & 97.32 & - & - & - & - \\
		\spacing LPN \cite{wang2021each} & 92.83 & 98.00 & 98.85 & 99.78 & 83.66 & 94.14 & 95.92 & 98.41 & - & - & - & -  \\
		\spacing TransGeo \cite{zhu2022transgeo} & 94.08 & 98.36 & 99.04 & 99.77 & 84.95 & 94.14 & 95.78 & 98.37 & - & - & - & - \\
		\spacing GeoDTR \cite{zhang2023cross} & 93.76 & 98.47 & 99.22 & 99.85 & 85.43 & 94.81 & 96.11 & 98.26 & 62.96 & 87.35 & 90.70 & 98.61 \\
		\spacing GeoDTR$^\dagger$ \cite{zhang2023cross} & 95.43 & 98.86 & 99.34 & 99.86 & 86.21 & 95.44 & 96.72 & 98.77 & 64.52 & 88.59 & 91.96 & 98.74 \\
		\spacing SAIG \cite{zhu2023simple} & 96.08 & 98.72 & 99.22 & 99.86 & 89.21 & 96.07 & 97.04 & 98.74 & - & - & - & - \\
		\spacing SAIG$^\dagger$ \cite{zhu2023simple} & 96.34 & 99.10 & 99.50 & 99.86 & 89.06 & 96.11 & 97.08 & \textbf{98.89} & 67.49 & 89.39 & 92.30 & 96.80 \\
		\spacing Sample4Geo \cite{deuser2023sample4geo} & 98.68 & 99.68 & 99.78 & \textbf{99.87} & 90.81 & \textbf{96.74} & \textbf{97.48} & 98.77 & 71.51 & 92.42 & 94.45 & 98.70 \\
		
		\spacing Baseline & 98.17 & 99.72 & 99.80 & 99.85 & 91.42 & 96.69 & 97.47 & 98.80 & 72.35 & 93.10 & \textbf{94.92} & \textbf{98.77} \\
		\spacing C-BEV (ours) & \textbf{98.72} & \textbf{99.77} & \textbf{99.83} & 99.85 & \textbf{91.68} & 96.62 & 97.42 & 98.75 & \textbf{75.94} & \textbf{93.23} & 94.85 & \textbf{98.77} \\
		
		\textbf{Unknown orientation} & & & & & & & & & & & & \\
		\spacing DSM \cite{shi2020looking} & 78.11 & 89.46 & 92.90 & 98.50 & 72.91 & 85.70 & 88.88 & 95.28 & - & - & - & - \\
		
		\spacing Baseline & 92.09 & 97.73 & 98.32 & 99.26 & 86.95 & 93.34 & 94.63 & 97.16 & 68.10 & 88.18 & 90.53 & \textbf{97.15} \\
		\spacing C-BEV (ours) & \textbf{97.13} & \textbf{99.07} & \textbf{99.16} & \textbf{99.29} & \textbf{89.46} & \textbf{94.62} & \textbf{95.46} & \textbf{97.17} & \textbf{73.94} & \textbf{90.53} & \textbf{92.17} & \textbf{97.15} \\
	\end{tabular}
\end{table*}

\begin{figure*}[!h]
	\centering
	\includegraphics[width=0.93\linewidth]{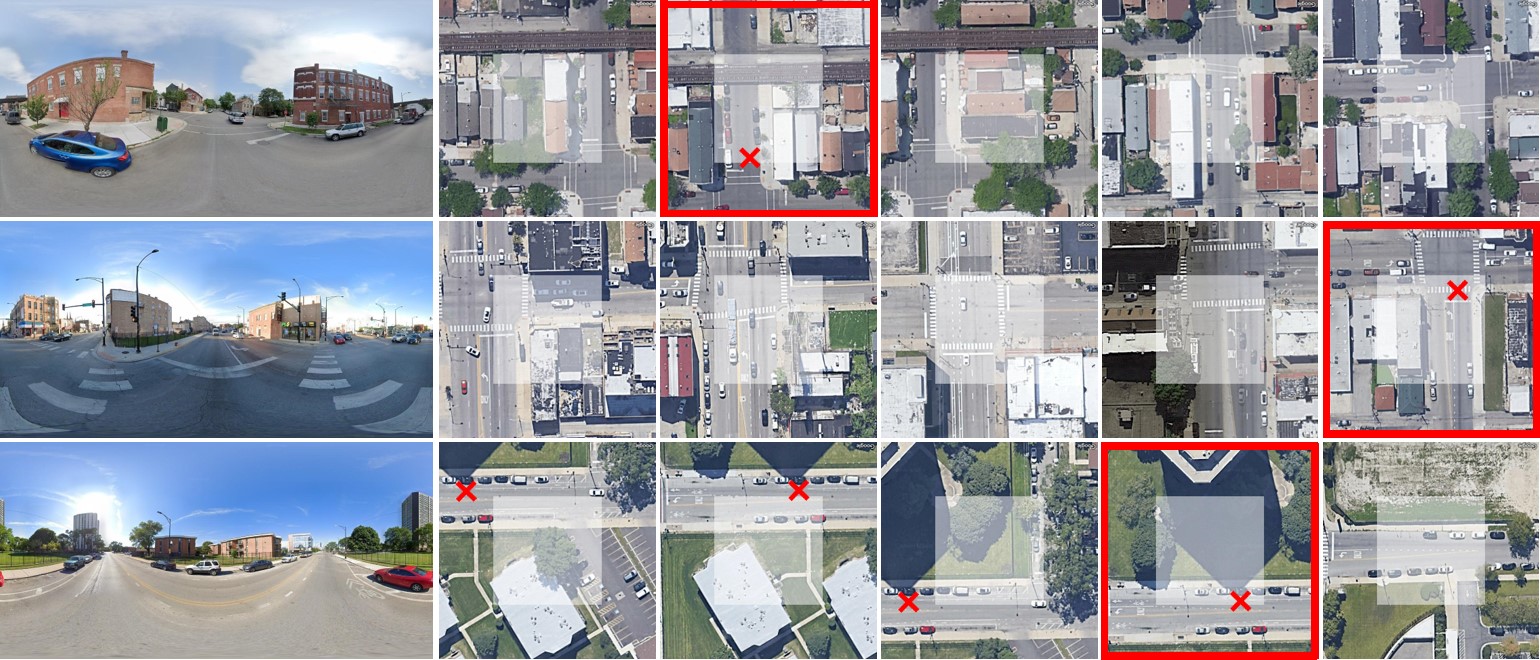}\vspace{-2mm}
	\caption{Examples of the BEV-based reranking on the VIGOR \cite{zhu2021vigor} dataset with unknown orientation. Each row shows a street-view query and the top-5 reference images retrieved using the first stage and ordered by decreasing matching score. A red border indicates the top-1 retrieved image after reranking in the second stage, and a red cross indicates the groundtruth position of the camera. White squares denote search regions per image.}
	\label{fig:retrieval-examples}
	\vspace{-4mm}
\end{figure*}

\subsection{Data}
\label{sec:data}

We conduct experiments on the CVUSA \cite{workman2015localize}, CVACT \cite{liu2019lending} and VIGOR \cite{zhu2021vigor} datasets which contain street-view panoramas with matching aerial images. We resize aerial images to shape $384 \times 384$ and panoramas to $384 \times 768$ in VIGOR and $140 \times 768$ in CVACT and CVUSA \cite{deuser2023sample4geo}.

VIGOR includes translational offsets between image pairs that correspond with a search region of half the image height and width. This allows (1) applying the trained model to real-world scenarios where a region is covered densely with (overlapping) aerial images as reference database, and (2) evaluating the performance of the metric 3-DoF pose estimation. In CVUSA and CVACT, aerial images are instead always centered on a preselected street-view query.

In all three datasets, street-view panoramas and aerial images are north-aligned by default. While most recent works evaluate only in this setting, we also provide results \wrt unknown orientation by randomly rotating (\ie horizontally shifting) the street-view panorama in the train and test splits by up to $360^\circ$. This demonstrates the feasibility of our method even in \mbox{real-world} scenarios where no external heading sensor is available. We decide against rotating the aerial images instead of the street-view panoramas \cite{zhu2021revisiting}, to avoid edges at the image corners or cropping (and resizing) the image to a different resolution.

The aerial images per dataset are provided with the same resolution in pixels, but cover different metric extents due to being sampled from a Mercator projection of the earth's surface. While all aerial images in CVACT are taken at a similar latitude with roughly $6.1 \frac{\cm}{\text{px}}$, images in VIGOR and CVUSA have a range of $10.1\frac{\cm}{\text{px}}$ to $11.8\frac{\cm}{\text{px}}$ and $12.7\frac{\cm}{\text{px}}$ to $27.2\frac{\cm}{\text{px}}$, respectively. 

For the first stage, we rescale aerial images to $384 \times 384$ resulting in varying metric image sizes. For the BEV-based retrieval which requires consistent metric resolutions, we resize images to $16.8\frac{\cm}{\text{px}}$ in VIGOR, $19.0\frac{\cm}{\text{px}}$ in CVACT and $53.0\frac{\cm}{\text{px}}$ in CVUSA, and crop or pad the result to $384 \times 384$ pixels. More detail on the datasets is provided in the supplemental material.

\subsection{Implementation details}

We use the ConvNeXt-B \cite{liu2022convnet} backbone as encoder for \mbox{street-view} and aerial images in both stages. We extract feature maps $F_P$ and $F_A$ at stride $8$ with $l_A = 48$. We sample $F_B$ at the same metric pixel size as $F_A$ with a sidelength of $l_B = 48$ for CVUSA and CVACT, and $l_B = 19$ for VIGOR following \cref{eq:lalb}. We use 256 channels for $F_P$ and $c = 32$ channels for $F_A$ and $F_B$. The search region is sampled with $n_t = 1$ translations for CVUSA and CVACT, $n_t = 28$ translations for VIGOR and $n_\theta = 32$ angles covering $360^\circ$ when evaluating with unknown orientation.

We train the first stage with a batch size of 128 for 40 epochs, update the embedding features every $e = 2$ epochs and use the AdamW \cite{loshchilov2017decoupled} optimizer with learning rate $1.0 \times 10^{-3}$, cosine decay and one epoch warmup. For the second stage, we change the batch size to 18 and train for 20 epochs with learning rate \mbox{$1.0 \times 10^{-4}$} instead. We use label smoothing of $0.1$ and a temperature of $\tau = 0.01$ in the loss function in both stages.

For data augmentation in the case of aligned orientation, we rotate both the aerial view and street view jointly by a random multiple of $90^\circ$. For unknown orientation, we rotate the street view by a random angle up to $360^\circ$ and the aerial view by a multiple of $90^\circ$, separately. During the first stage, we randomly flip pairs of street-view and aerial images horizontally. At test time, we retrieve $k=100$ reference images per query from the first stage and rerank them using the second stage.

\subsection{Image retrieval}

\begin{table}[t]
	\setlength{\tabcolsep}{6pt} %
	\newcommand{\spacing}{\hspace{3mm}}
	\small
	\centering
	\caption{Ablation studies for the image retrieval on the VIGOR dataset with known orientation. Our baseline is the first stage of our method and trained similar to Sample4Geo \cite{deuser2023sample4geo}.}
	\label{tab:ablation-retrieval}
	\vspace{-2mm}
	\begin{tabular}{l|cc|cc}
		& \multicolumn{2}{c|}{Same-area} & \multicolumn{2}{c}{Cross-area} \\
		& R@1 & R@5 & R@1 & R@5 \\
		\hline
		\spacing Baseline & 78.25 & 95.70 & 61.26 & 82.49 \\
		\textbf{BEV resolution} & & & & \\
		\spacing Stride 4  & 86.69 & 97.50 & 78.67 & 91.59 \\
		\spacing Stride 8 & \textbf{87.49} & \textbf{97.65} & 80.01 & \textbf{92.02} \\
		\spacing Stride 16 & 86.85 & 97.59 & \textbf{80.27} & 91.80 \\
		\textbf{First stage as prior} & & & & \\
		\spacing With & \textbf{87.49} & \textbf{97.65} & \textbf{80.01} & \textbf{92.02} \\
		\spacing Without & 84.49 & 95.62 & 76.77 & 90.41 \\
		\textbf{Number of candidates} & & & & \\
		\spacing $k=10$ & 87.16 & 96.72 & 77.01 & 86.42 \\
		\spacing $k=100$ & 87.49 & 97.65 & 80.01 & 92.02 \\
		\spacing $k=1000$ & \textbf{87.54} & \textbf{97.78} & \textbf{80.31} & \textbf{92.83} \\
	\end{tabular}
	\vspace{-3mm}
\end{table}

\begin{figure}[t]
	\centering
	\includegraphics[width=\linewidth]{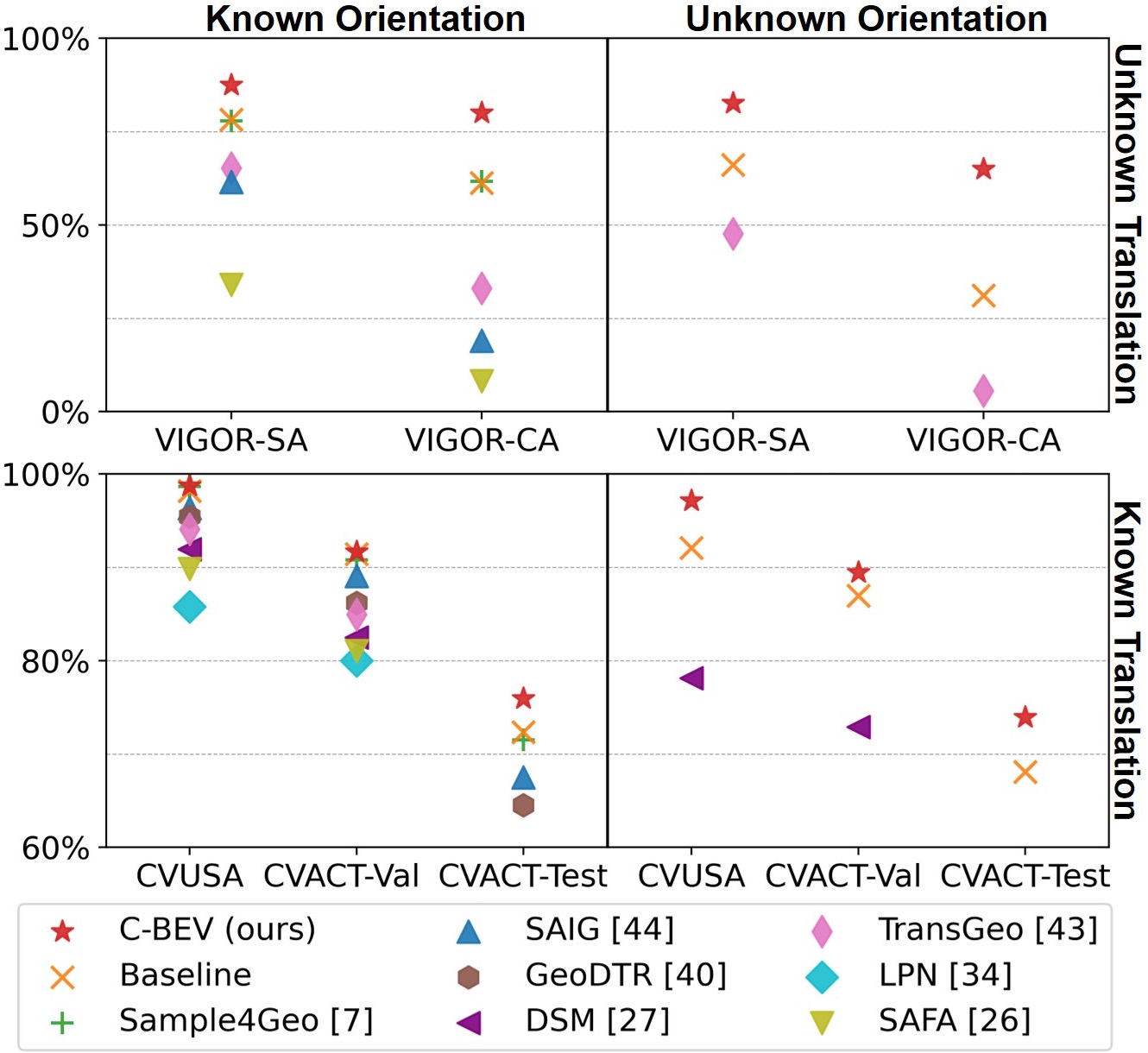}\vspace{-2mm}
	\caption{Top-1 recall of our retrieval method C-BEV ($\vcenter{\hbox{\includegraphics[height=2.0mm]{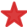}}\vspace{0.4mm}}$) compared with previous approaches. Our work pushes the state-of-the-art by a large margin, especially in challenging real-world settings where aerial and street-view images are not preselected to align \wrt translation and orientation. We train our baseline ($\vcenter{\hbox{\includegraphics[height=2.0mm]{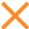}}\vspace{0.5mm}}$) similar to Sample4Geo \cite{deuser2023sample4geo} ($\vcenter{\hbox{\includegraphics[height=2.0mm]{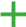}}\vspace{0.5mm}}$). Most recent works consider only \mbox{north-aligned} images, \ie with known orientation.}
	\label{fig:summary-results}
	\vspace{-4mm}
\end{figure}

\paragraph{Setting} We evalute the image retrieval performance of our method \mbox{C-BEV} on the VIGOR (\cf \cref{tab:vigor-recall}), CVUSA and CVACT (\cf \cref{tab:cv-recall}) datasets, each with known and unknown orientation, and report the top-k recall after the first and second stage (\cf \cref{fig:summary-results}).

\paragraph{Results} Our method outperforms previous works in all evaluated many-to-one scenarios (\ie where the relative translation or orientation between street-view and aerial images is unknown), and achieves comparable or better performance in one-to-one scenarios (\ie where aerial images are preselected to align with street-view images \wrt both translation and orientation). The improvement is particularly large in more challenging settings, \eg increasing the top-1 recall on the VIGOR splits by $9.2$ and $18.8$ percentage points with aligned orientation and $16.5$ and $33.9$ points with unknown orientation. \cref{fig:retrieval-examples} shows examples of the \mbox{BEV-based} reranking on the VIGOR dataset.

\begin{table*}[t]
	\setlength{\tabcolsep}{3pt} %
	\newcommand{\spacing}{\hspace{3mm}}
	\small
	\centering
	\caption{Mean and median translation and orientation error on matching image pairs in the VIGOR dataset. *Unlike previous works, our method is trained without metric supervision and only requires pairwise assignment of images as groundtruth.}
	\vspace{-2mm}
	\label{tab:vigor-metric}
	\begin{tabular}{l|cc|cc|cccc|cccc}
		& \multicolumn{4}{c|}{Known orientation} & \multicolumn{8}{c}{Unknown orientation} \\
		& \multicolumn{2}{c|}{Same-area} & \multicolumn{2}{c|}{Cross-area} & \multicolumn{4}{c|}{Same-area} & \multicolumn{4}{c}{Cross-area} \\
		& Mean & Median & Mean & Median & \multicolumn{2}{c}{Mean} & \multicolumn{2}{c|}{Median} & \multicolumn{2}{c}{Mean} & \multicolumn{2}{c}{Median} \\
		\hline
		\textbf{Supervised} & & & & & & & & & & & & \\
		\spacing CVR \cite{zhu2021vigor} & 8.99m & 7.81m & 8.89m & 7.73m & - & - & - & - & - & - & - & - \\
		\spacing MCC \cite{xia2022visual} & 6.94m & 3.64m & 9.05m & 5.14m & 9.87m & 56.86$^\circ$ & 6.25m & 16.02$^\circ$ & 12.66m & 72.13$^\circ$ & 9.55m & 29.97$^\circ$ \\
		\spacing SliceMatch \cite{lentsch2023slicematch} & 5.18m & 2.58m & 5.53m & 2.55m & 6.49m & 25.46$^\circ$ & 3.13m & 4.71$^\circ$ & 7.22m & 25.97$^\circ$ & \textbf{3.31m} & 4.51$^\circ$ \\
		\spacing GGCVT \cite{shi2023boosting} & 4.12m & \textbf{1.34m} & 5.16m & \textbf{1.40m} & - & - & - & - & - & - & - & - \\
		\textbf{Unsupervised*} & & & & & & & & & & & & \\
		\spacing C-BEV (ours) & \textbf{3.52m} & 2.58m & \textbf{3.97m} & 2.86m & \textbf{3.85m} & \textbf{5.55$^\circ$} & \textbf{2.81m} & \textbf{2.16$^\circ$} & \textbf{4.78m} & \textbf{7.90$^\circ$} & 3.48m & \textbf{2.52$^\circ$} \\
	\end{tabular}
	\vspace{-5mm}
\end{table*}

\paragraph{Ablation study on BEV resolution} We test for suitable resolutions of the BEV maps by encoding $F_A$ and $F_B$ at strides 4, 8 and 16 (\ie $0.67\frac{\m}{\text{px}}$, $1.34\frac{\m}{\text{px}}$ and $2.68\frac{\m}{\text{px}}$) with 128, 32 and 8 channels, respectively. The resolutions are chosen such that the BEV maps have the same total size, but constitute different trade-offs between spatial resolution and representational capacity per cell. While the difference between resolutions is small, the best results are achieved at stride 8 (\cf \cref{tab:ablation-retrieval}).

\paragraph{Ablation study on first stage as soft prior} To examine the impact of including the first stage's scores as prior probabilities in the second stage's reranking, we evaluate the retrieval performance after reranking with the BEV-based scores only (\cf \cref{tab:ablation-retrieval}). The recall drops slightly, indicating that the BEV-based reranking is suited most for discriminating hard samples and achieves best performance in combination with vector-based retrieval that provides a sufficient prior.

\paragraph{Ablation study on number of candidates} \Cref{tab:ablation-retrieval} shows that reducing the number of candidates per query for the second stage from $100$ to $10$ results in significantly lower recall on the VIGOR cross-area split. At $1000$ candidates the computation time is increased significantly, but yields only small improvement. We choose $k = 100$ candidates as a trade-off between runtime and recall.

\paragraph{Runtime and scalability} We evaluate the runtime of the retrieval on the VIGOR dataset using an Nvidia H100. Our implementation tests multiple reference images per query in parallel. The computation of embedding maps $F_A$ and $F_B$ requires about $1.3\ms$ and $4.6\ms$, respectively. The matching operation requires about $0.1 \ms$ to $0.3\ms$ per candidate, dependent on the choice of $n_\theta$ up to 32.

We rerank 100 candidates per query image in the VIGOR dataset, \ie about $0.1\%$ of the possible reference images each covering a search region of about $35\m \times 35\m$. Given an area that is densely covered with aerial images in a similar setup, the reranking incurs an additional cost of \mbox{$(\frac{1000\m}{35\m})^2 \cdot 0.1\% \approx 1$} matching operation per square kilometer. Without precomputed BEV embeddings, this corresponds to roughly $1.4\ms$ to $1.6\ms$ per square kilometer, and about 4 hours for the land area of the United States.

\subsection{Pose estimation} \label{sec:eval-pose}

\paragraph{Setting} We evaluate the pose estimation performance of \mbox{C-BEV} on matching image pairs in the VIGOR dataset \mbox{(\cf \cref{tab:vigor-metric})} with known and unknown orientation and report the mean and median translation and orientation error.

\begin{figure}
	\centering
	\vspace{1mm}\includegraphics[width=\linewidth]{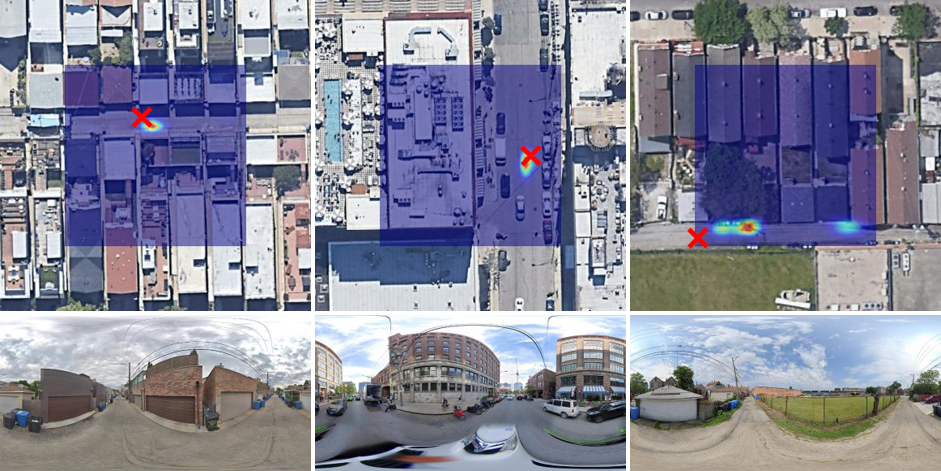}\vspace{-1mm}
	\caption{Examples of probability distributions over camera locations predicted by our method. The model learns to infer the camera pose despite not receiving metric supervision during training. Search regions are larger than half the image diameter due to the cropping of images described in \cref{sec:data}.}
	\label{fig:probs}
	\vspace{-3mm}
\end{figure}

\paragraph{Results} Our method achieves a better mean pose error than recent pose estimation methods on VIGOR in all settings despite training without any metric groundtruth. \cref{fig:probs} shows examples of the predicted probability distributions over possible camera locations in the search region.

The latest supervised work GGCVT \cite{shi2023boosting} achieves a smaller median position error for the case of aligned orientation, but larger mean error than our method. The gap between mean and median metrics indicates that C-BEV is better at providing a rough camera pose estimate, but does not achieve the same fine accuracy as GGCVT.

\paragraph{Ablation study on search region boundary} To further examine the model's ability to infer accurate camera locations without explicit supervision, we consider the case of \mbox{street-view} images being located close to the search region boundary. Here, the loss might provide direct metric supervision by penalizing small offsets of the predicted camera location across the search region boundary onto a negative reference image. To test this hypothesis, we train a model by excluding negative reference images per batch that are located closer than 50m to a \mbox{street-view} query. The resulting model yields only marginally worse pose predictions, which suggests that boundary cases are not the primary cause of the model's ability to infer camera locations without direct supervision.

\section{Conclusion}
\label{sec:conclusion}

We introduce our method C-BEV for the cross-view geo-localization task. Our approach uses a novel embedding representation in the form of bird's eye view maps, and specifically addresses the many-to-one ambiguity that arises when varying street-view camera poses with vastly differing visual appearances have to be matched to the same aerial image.

C-BEV surpasses the state-of-the-art on the cross-view image retrieval task by a large margin, especially in challenging real-world settings. Even though we use only a contrastive objective applied on image pairings as supervision during training, our model learns to infer the 3-DoF camera pose on the matching aerial image, and even yields a better mean pose error than recent methods explicitly trained with metric groundtruth.

{\small
\bibliographystyle{ieeenat_fullname}
\bibliography{11_references}
}

\ifarxiv \clearpage \appendix {
	\centering
	\Large
	\textbf{Appendix}
	\vspace{0.4em}
}

\section{Overview of datasets}

\Cref{fig:dataset-maps} gives an overview of the datasets used for evaluation in this paper. In VIGOR, aerial images cover several regions densely (\cf \cref*{sec:problem-formulation}), and up to two street-view queries are chosen in the search region of each reference image. In CVUSA and CVACT, aerial images are preselected to align with a set of street-view images.

All aerial images per dataset are sampled with the same size in pixels from a Mercator projection of the earth's surface. As a consequence, the size in meters \mbox{(\cf \cref{tab:datasets-metadata})} changes with the image's latitude (\ie distance north or south of the equator).

\begin{figure*}
	\centering
	\begin{subfigure}[t]{0.91\textwidth}
		\centering
		\includegraphics[width=\textwidth]{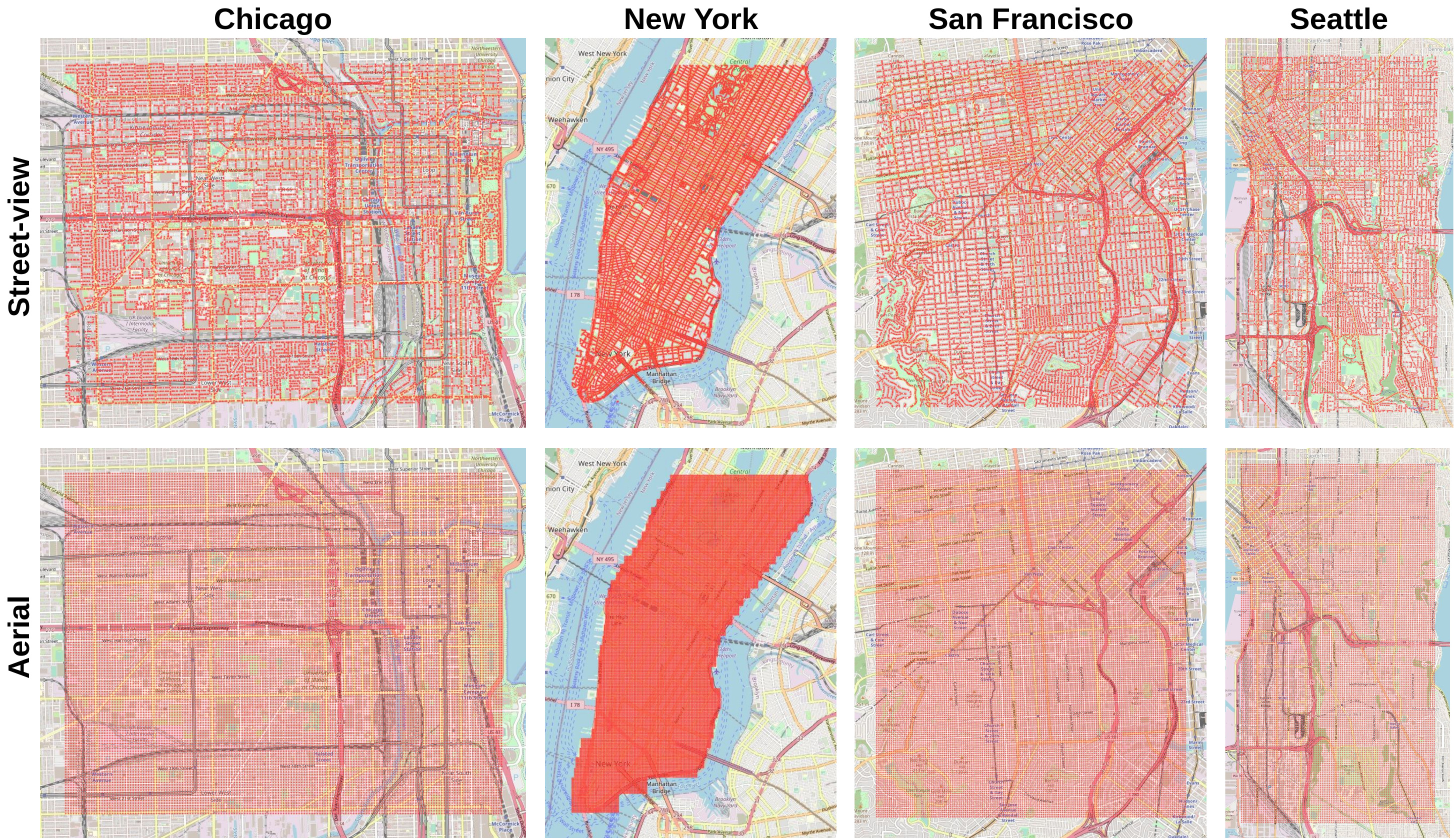}
		\caption{Images in VIGOR \cite{zhu2021vigor} are captured in four different cities in the United States. Every dot represents the location  of an image. Aerial images cover a region in a regular grid while street-view images are sampled at random locations.\vspace{3mm}}
		\label{fig:maps-vigor}
	\end{subfigure}
	\begin{subfigure}[t]{0.91\textwidth}
		\centering
		\includegraphics[width=\linewidth]{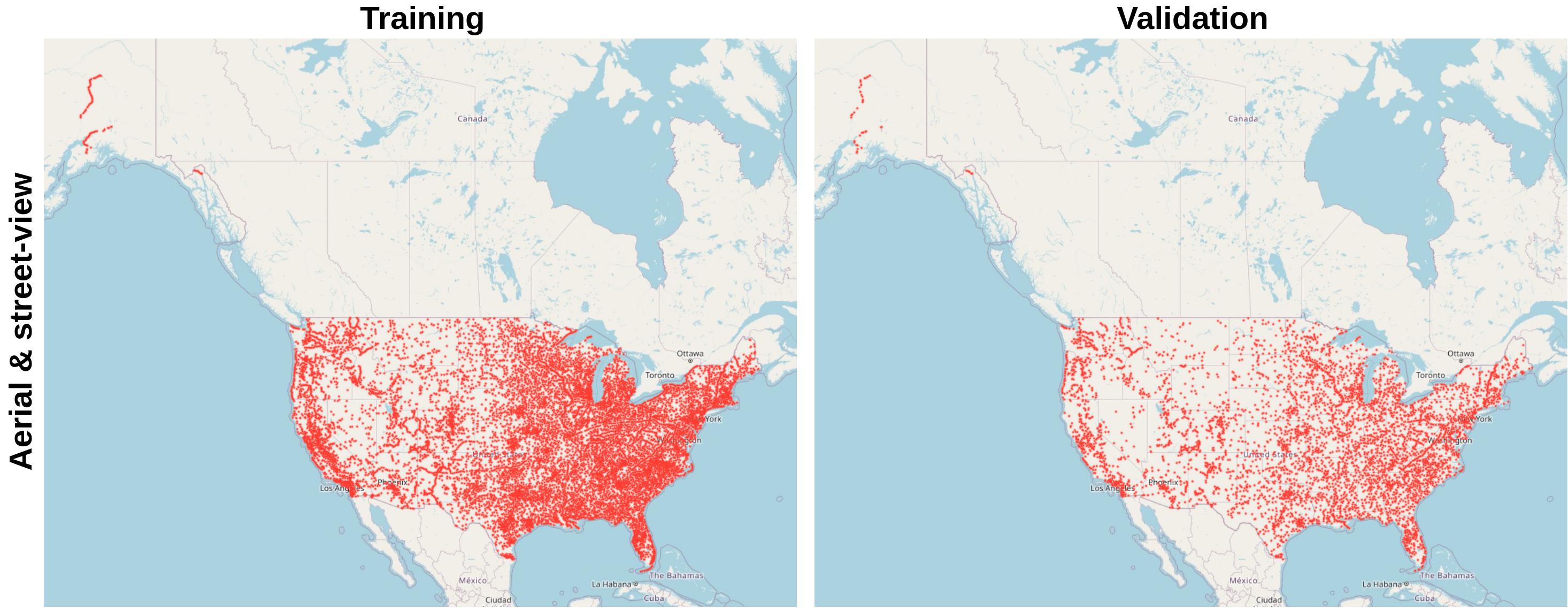}
		\caption{Images in CVUSA \cite{workman2015wide} are captured in the United States. Every dot represents a pair of aligned street-view and aerial images.\vspace{3mm}}
		\label{fig:maps-cvusa}
	\end{subfigure}
	\begin{subfigure}[t]{0.91\textwidth}
		\centering
		\includegraphics[width=\linewidth]{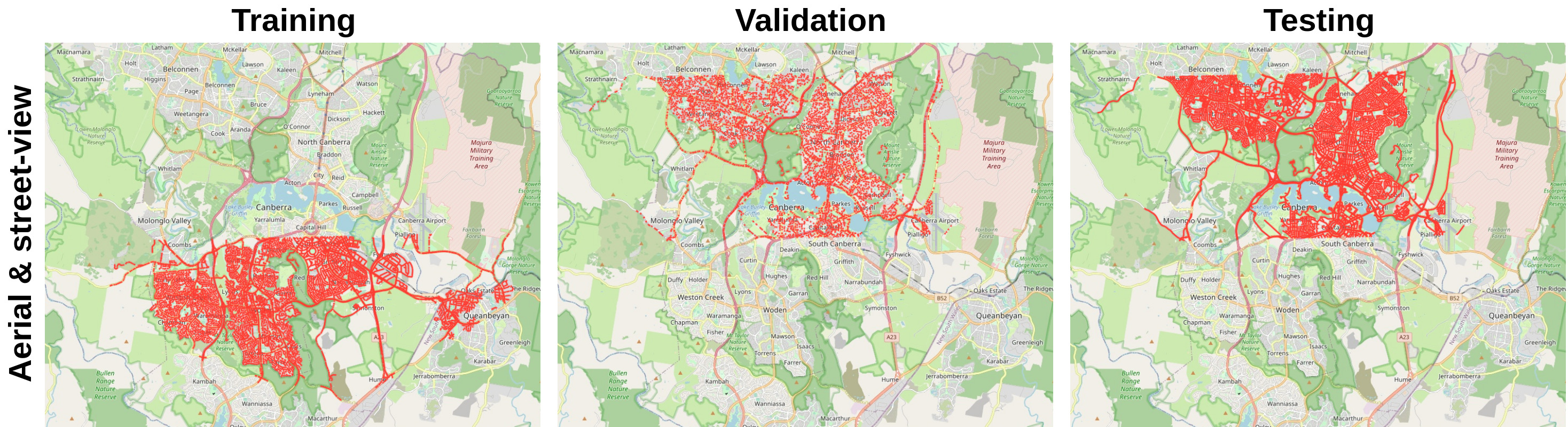}
		\caption{Images in CVACT \cite{liu2019lending} are captured in the Australian Capital Territory. Every dot represents a pair of aligned street-view and aerial images.}
	\end{subfigure}
	\caption{Overview of image locations in datasets used in this work. Map data from \href{https://www.openstreetmap.org/copyright}{OpenStreetMap} \cite{OpenStreetMap}.\label{fig:dataset-maps}}
\end{figure*}

\section{Effect of metric pixel size}

\paragraph{Setting} We report results for cross-dataset image retrieval (\cf \cref{tab:cv-recall-crossdataset}) where the models trained on CVUSA and CVACT are also tested on each other's validation splits as proposed by Yang \etal \cite{yang2021cross}, denoted as \mbox{CVUSA $\rightarrow$ CVACT} and \mbox{CVACT $\rightarrow$ CVUSA}.

\paragraph{Effect of metric pixel size} For our baseline using vector embeddings, we rescale all aerial images with the same factor per dataset to $384 \times 384$ pixels following related works and achieve similar results as the current state-of-the-art method Sample4Geo \cite{deuser2023sample4geo}. Resizing the images to a consistent pixel size of $19.0\frac{\text{cm}}{\text{px}}$ already improves the baseline results significantly, even though a large portion of the images in CVUSA are cropped as a consequence. This indicates that the model does not generalize well to aerial images with unseen scale. However, such a capability is not required in the context of cross-view geolocalization:  Aerial images are always georeferenced and their scale is known, such that they can be resized as required.

Of the datasets used in this work (\cf \cref{tab:cv-recall-crossdataset}), we observe the impact of resizing aerial images to consistent metric pixel sizes mainly in the cross-dataset setting. \Eg in \mbox{CVACT $\rightarrow$ CVUSA}, models are trained on images with sidelength 73m and tested on 95m to 204m. While the CVUSA dataset also contains images with varying pixel sizes, this range is the same in both train and test splits, such that all scales are also seen during training. In this case, resizing images to the same metric pixel size (with corresponding cropping or padding to $384 \times 384$ pixels) degrades the performance, potentially due to the loss of information incurred from cropping or using coarser pixel resolutions.

Our evaluation demonstrates that training and testing on the same range of metric pixel sizes is preferred. We thus report our main results in \cref*{sec:evaluation} on splits where the range of pixel sizes of testing images is contained in or close to the range of the training images.

\begin{table}
	\setlength{\tabcolsep}{6pt} %
	\newcommand{\spacing}{\hspace{3mm}}
	\small
	\centering
	\caption{Sidelength (in meters) of aerial images in datasets used in this work. Each row corresponds to a different train-test split.\vspace{-1mm}}
	\label{tab:datasets-metadata}
	\begin{tabular}{l|rr|rr}
		& \multicolumn{2}{c|}{Training images} & \multicolumn{2}{c}{Testing images} \\
		& Min. & Max. & Min. & Max. \\
		\hline
		VIGOR same-area \cite{zhu2021vigor} & 64.3 & 75.3 & 64.3 & 75.3 \\
		VIGOR cross-area \cite{zhu2021vigor} & 64.3 & 72.3 & 71.0 & 75.3 \\
		CVUSA \cite{workman2015wide} & 95.1 & 203.7 & 95.2 & 203.5 \\
		CVACT-Val \cite{liu2019lending} & 72.8 & 72.8 & 72.8 & 72.9 \\
		CVACT-Test \cite{liu2019lending} & 72.8 & 72.8 & 72.8 & 72.9 \\
		CVUSA $\rightarrow$ CVACT & 95.1 & 203.7 & 72.8 & 72.9 \\
		CVACT $\rightarrow$ CVUSA & 72.8 & 72.8 & 95.2 & 203.5 \\
	\end{tabular}
\end{table}

\paragraph{Results of our method} We report results of C-BEV in the cross-dataset setting with images resized to $19.0\frac{\text{cm}}{\text{px}}$ (and cropped to $384 \times 384$ pixels) as it requires consistent pixel sizes. Since related works evaluate with the original metric sizes, we train a model similar to the best previous work Sample4Geo \cite{deuser2023sample4geo} using resized images as our baseline.

Our method improves the results significantly, especially in challenging settings where current state-of-the-art approaches do not perform well. In the case of unknown orientation, the top-1 recall is improved by $22.7$ and $9.2$ percentage points over the baseline.

\section{Effect of search region boundary}

We evaluate the 3-DoF pose estimation performance of our method on the VIGOR dataset with a minimum distance between street-view images and negative aerial reference images per batch to examine the impact of query images being located close to the search region boundary as discussed in \cref*{sec:eval-pose}. \Cref{tab:vigor-metric-ablation} shows that the recall drops only slightly when including an offset of $50\text{m}$, indicating that search region boundaries are not primarily responsible for the ability of the model to infer the 3-DoF camera pose without explicit supervision.

\begin{table*}
	\setlength{\tabcolsep}{5pt} %
	\newcommand{\spacing}{\hspace{3mm}}
	\small
	\centering
	\caption{Top-k recall in a cross-dataset setting as proposed by Yang \etal \cite{yang2021cross}: Models are trained on the CVUSA and CVACT training splits and tested on each other's validation splits. Our baseline is the first stage of our method and trained similar to Sample4Geo \cite{deuser2023sample4geo}. $\dagger$ denotes methods that use polar transformation on the aerial input image. * indicates that images in CVUSA and CVACT are resized to the same metric size with $19.0\frac{\text{cm}}{\text{px}}$.}
	\label{tab:cv-recall-crossdataset}
	\begin{tabular}{l|cccc|cccc}
		& \multicolumn{4}{c|}{CVUSA $\rightarrow$ CVACT} & \multicolumn{4}{c}{CVACT $\rightarrow$ CVUSA} \\
		& R@1 & R@5 & R@10 & R@1\% & R@1 & R@5 & R@10 & R@1\% \\
		\hline
		\textbf{Known orientation} & & & & & & & & \\
		\spacing L2LTR$\dagger$ \cite{yang2021cross} & 47.55 & 70.58 & 77.39 & 91.39 & 33.00 & 51.87 & 60.63 & 84.79 \\
		\spacing GeoDTR$\dagger$ \cite{zhang2023cross} & 53.16 & 75.62 & 81.90 & 93.80 & 44.07 & 64.66 & 72.08 & 90.09 \\
		\spacing Sample4Geo \cite{deuser2023sample4geo} & 56.62 & 77.79 & 87.02 & 94.69 & 44.95 & 64.36 & 72.10 & 90.65\\
		
		\spacing Baseline & 58.73 & 82.35 & 87.65 & 96.41 & 42.72 & 64.99 & 72.97 & 92.45 \\
		\hdashline
		\spacing Baseline* & 85.02 & 95.50 & 96.68 & \textbf{98.47} & 64.77 & 79.91 & 84.85 & 96.09 \\
		\spacing C-BEV* (ours) & \textbf{88.07} & \textbf{96.15} & \textbf{97.07} & 98.46 & \textbf{70.01} & \textbf{83.57} & \textbf{87.82} & \textbf{96.25} \\
		\hline
		\textbf{Unknown orientation} & & & & & & & & \\
		
		\spacing Baseline & 25.87 & 45.99 & 54.66 & 79.11 & 12.70 & 24.30 & 30.55 & 55.14 \\
		\hdashline
		\spacing Baseline* & 55.32 & 76.18 & 81.70 & 92.91 & 35.49 & 49.47 & 55.39 & 74.58 \\
		\spacing C-BEV* (ours) & \textbf{78.06} & \textbf{88.63} & \textbf{90.30} & \textbf{93.37} & \textbf{44.71} & \textbf{56.80} & \textbf{61.89} & \textbf{75.37} \\
	\end{tabular}
\end{table*}

\begin{table*}
	\setlength{\tabcolsep}{4pt} %
	\newcommand{\spacing}{\hspace{3mm}}
	\small
	\centering
	\caption{Mean and median translation and orientation error of our method C-BEV on matching image pairs in the VIGOR dataset. We add a minimum distance between street-view queries and negative reference images per batch in the data sampler during training.}
	\label{tab:vigor-metric-ablation}
	\begin{tabular}{c|cc|cc|cccc|cccc}
		& \multicolumn{4}{c|}{Known orientation} & \multicolumn{8}{c}{Unknown orientation} \\
		& \multicolumn{2}{c|}{Same-area} & \multicolumn{2}{c|}{Cross-area} & \multicolumn{4}{c|}{Same-area} & \multicolumn{4}{c}{Cross-area} \\
		Min. distance & Mean & Median & Mean & Median & \multicolumn{2}{c}{Mean} & \multicolumn{2}{c|}{Median} & \multicolumn{2}{c}{Mean} & \multicolumn{2}{c}{Median} \\
		\hline
		0m & 3.52m & 2.58m & 3.97m & 2.86m & 3.87m & 5.56$^\circ$ & 2.81m & 2.16$^\circ$ & 4.78m & 7.90$^\circ$ & 3.48m & 2.52$^\circ$ \\
		
		50m & 3.53m & 2.70m & 4.21m & 3.31m & 3.82m & 5.64$^\circ$ & 2.81m & 2.20$^\circ$ & 5.04m & 8.24$^\circ$ & 3.70m & 2.66$^\circ$ \\
	\end{tabular}

\end{table*}

 \fi

\end{document}